\documentclass[11pt,a4paper]{article}
\usepackage[hyperref]{eacl2021}
\usepackage{times}
\usepackage{latexsym}

\usepackage[linesnumbered,ruled,vlined]{algorithm2e}
\usepackage{bm}
\usepackage{amsmath}
\usepackage{mathtools}

\usepackage{subcaption}
\usepackage{booktabs}
\usepackage{amsfonts}
\usepackage{xcolor}
\usepackage{multirow}

\usepackage{microtype}

\aclfinalcopy % Uncomment this line for the final submission
\def\aclpaperid{1037} %  Enter the acl Paper ID here

\title{Complementary Evidence Identification in  \\
Open-Domain Question Answering}

\author{
    Xiangyang Mou \\
    Rensselaer Polytechnic Institute \\
    \texttt{moux4@rpi.edu}
\\\And
    Mo Yu \\
    IBM Research  \\
    \texttt{yum@us.ibm.com}
\\\And
    Shiyu Chang \\
    MIT-IBM Watson AI Lab  \\
    \texttt{shiyu.chang@ibm.com}
\\\AND
    Yufei Feng \\
    Queen’s University  \\
    \texttt{feng.yufei@queensu.ca}
\\\And
    Li Zhang  \\
    Amazon Web Services  \\
    \texttt{lzhangza@amazon.com}
\\\And
    Hui Su \\
    Fidelity \\
    \texttt{Hui.Su@fmr.com}
\\}

\date{}

\begin{document}
\maketitle
\begin{abstract}
This paper proposes a new problem of complementary evidence identification for open-domain question answering (QA).  The problem aims to efficiently find a small set of passages that covers full evidence from multiple aspects as to answer a complex question.  To this end, we proposes a method that learns vector representations of passages and models the sufficiency and diversity within the selected set, in addition to the relevance between the question and passages. Our experiments demonstrate that our method considers the dependence within the supporting evidence and significantly improves the accuracy of complementary evidence selection in QA domain.

%%%%%%  VERSION 1  %%%%%%  
% This paper proposes a new problem of complementary evidence identification for open-domain question answering.  The problem aims to efficiently find a small set of paragraphs that covers full evidence for answering a complex question that asks for multiple facts.  To this end, in addition to the relevance between a question and passages, we define the sufficiency and diversity of the selected set. A method is further proposed that learns vector representations of passages and models the above criterion.  Experiments on the multi-hop QA dataset HotpotQA demonstrate that our method considers the dependence within the supporting evidence and significantly improves the accuracy of complementary evidence selection.

\end{abstract}

\section{Introduction}

\begin{figure}
    \centering
    \includegraphics[width=0.48\textwidth]{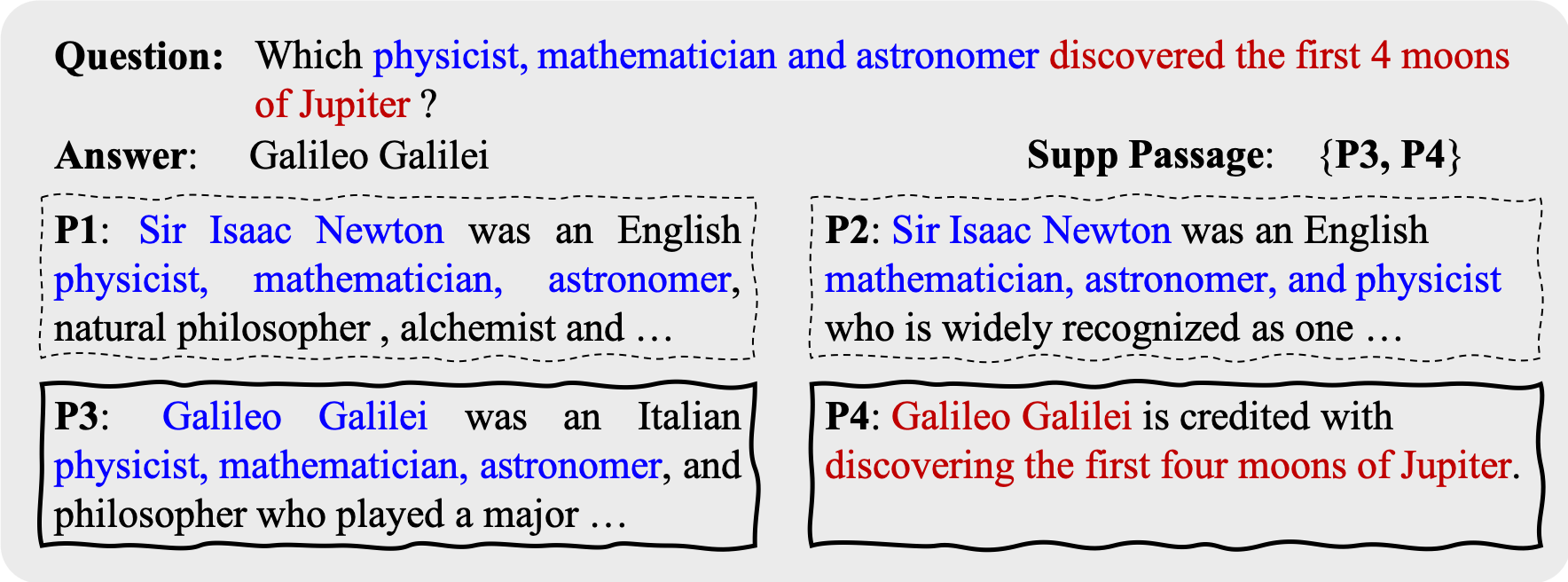}
    \includegraphics[width=0.48\textwidth]{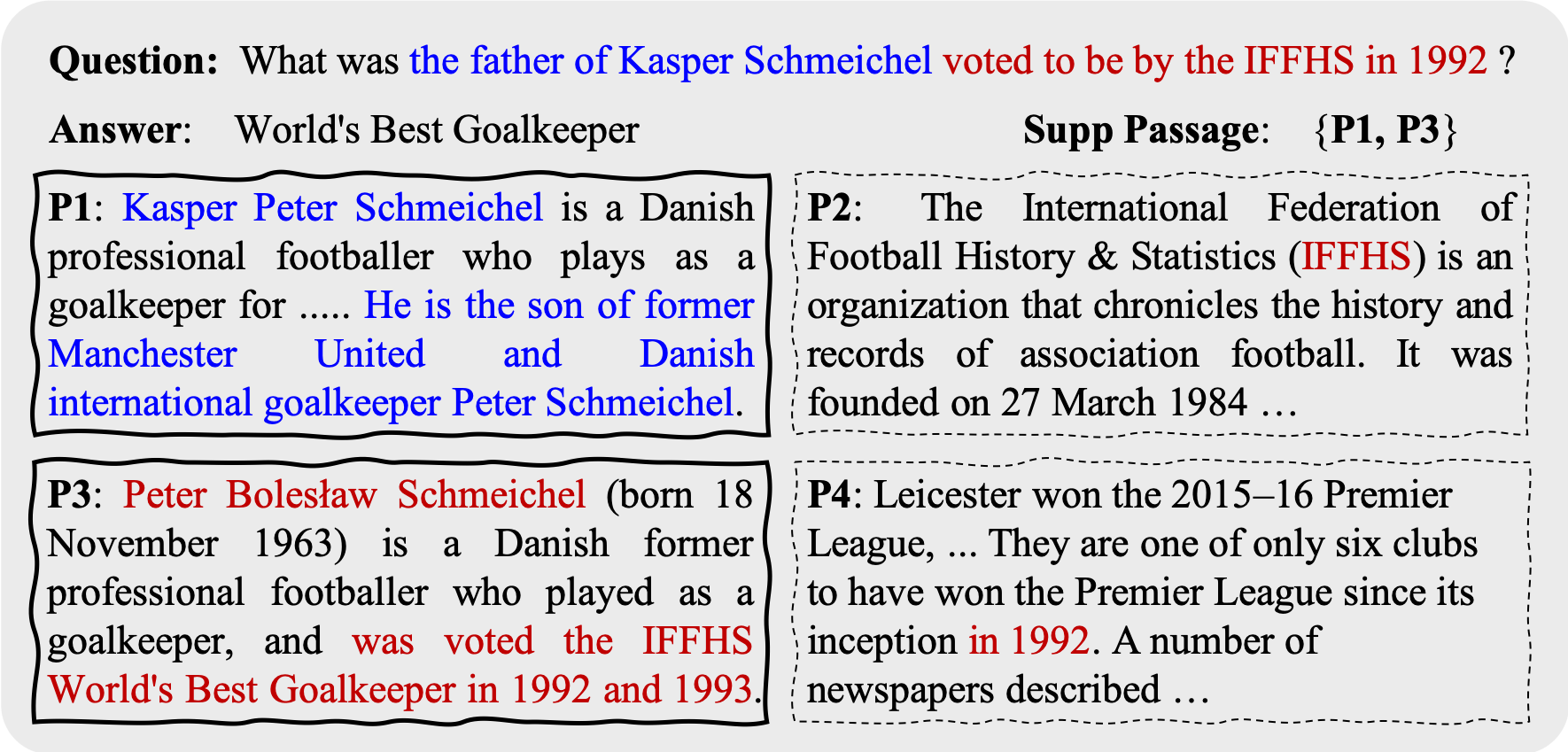}
    \caption{Examples of complex questions involving two facts of a person. Different facts are color-coded. \textbf{P\#} are all relevant passages, while only the ones with solid-line boxes are the true supporting passages.}
    \label{fig:multi_hop_example}
\end{figure}

In recent years, significant progress has been made in the field of open-domain question answering ~\cite{chen2017reading,wang2017evidence,wang2018r3,clark2018simple,min2018efficient,asai2019learning}.  Very recently, some works turn to deal with a more challenging task of asking complex questions~\cite{welbl2018constructing,clark2018think,yang2018hotpotqa} from the open-domain text corpus.  In the open-domain scenario, one critical challenge raised by complex questions is that each question may require multiple pieces of evidence to get the right answer, while the evidence usually scatters in different passages.  Examples in Figure~\ref{fig:multi_hop_example} shows two types of questions that require evidence from multiple passages.

To deal with the challenging multi-evidence questions, an open-domain QA system should be able to (1) efficiently retrieve a small number of passages that cover the full evidence; and (2) accurately extract the answer by jointly considering the candidate evidence passages. While there have been several prior works in the latter direction~\cite{wang2017evidence,clark2018simple,lin2018denoising}, the solutions to the first problem still rely on traditional or neural information retrieval (IR) approaches, which solely measure the relevance between the question and each individual paragraph, and will highly possibly put the wrong evidence to the top.\footnote{\cite{min2019compositional} pointed out the shortcut problem in multi-hop QA. However, as some works~\cite{wang2019multi} show that even a better designed multi-hop model can still benefit from full evidence in such situation.} For example in Figure~\ref{fig:multi_hop_example} (top), \textbf{P1} and \textbf{P2} are two candidate evidence passages that are closely related to the question but only cover the same unilateral fact required by the question, therefore leading us to the wrong answer \emph{Newton}.

This paper formulates a new problem of \textbf{complementary evidence identification} for answering complex questions. The key idea is to consider the problem as measuring the properties of the selected passages, more than the individual relevance. Specifically, we hope the selected passages can serve as a set of spanning bases that supports the question. The selected passage set thus should satisfy the properties of (1)\emph{relevancy}, i.e., they should be closely related to the question; (2) \emph{diversity}, i.e., they should cover diverse information given the coverage property is satisfied; (3) \emph{compactness}, i.e., the number of passages to satisfy the above properties should be minimal. With these three defined properties, we hope to both improve the selective accuracy and encourage the interpretability of the evidence identification. Note that complementary evidence identification in QA is different from Search Result Diversification (SRD) in IR on their requirement of compactness. The size of the selected set is constrained in QA tasks by the capability of downstream reasoning models and practically needs to be a small value, whereas it is not the case in SRD.

To achieve the above goals, a straightforward approach is to train a model that evaluates each subset of the candidate passages, e.g., by concatenating passages in any subsets. However, this approach is highly inefficient since it requires to encode $O(K^L)$ passage subsets, where $K$ is the total number of candidates and $L$ is the maximum size of subsets.  Thus, a practical complementary evidence identification method needs to be computationally efficient. This is especially critical when we use heavy models like ELMo~\cite{peters2018deep} and BERT~\cite{devlin2018bert}, where passage encoding is time and memory consuming.

To this end, we propose an efficient method to select a set of spanning passages that is sufficient and diverse.  The core idea is to represent questions and passages in a vector space and define the measures of our criterion in the vector space. For example, in the vector space, sufficiency can be defined as a similarity between the question vector and the sum of selected passage vectors, measured by a cosine function with a higher score indicating a closer similarity; and diversity can be defined as $\ell_1$ distance between each pair of passages.  By properly training the passage encoder with a loss function derived by the above terms, we expect the resulted vector space satisfies the property that the complementary evidence passages lead to large scores. In addition, our method only encodes each passage in the candidate set once, which is more efficient than the naive solution mentioned above.  To evaluate the proposed method, we use the multi-hop QA dataset HotpotQA (the full wiki setting) since the ground-truth of evidence passages are provided. Experiments show that our method significantly improves the accuracy of complementary evidence selection.
%%%%% Version 1 %%%%% 
% To this end, we propose an efficient method to select a set of spanning passages that is sufficient and diverse.  The core idea is to represent questions and passages in a vector space and define the measures of our criterion in the vector space. For example, in the vector space, sufficiency can be defined as a similarity between the question vector and the summation of selected passage vectors, measured by a cosine function; and diversity can be defined as $\ell_1$ distance between each pair of passages.  By properly training the passage encoder with a loss function derived by the above terms, we expect the resulted vector space satisfies the property that the complementary evidence passages lead to small losses. In addition, our method only encodes each passage in the candidate set once, which is more efficient than the naive solution mentioned above.  To evaluate the proposed method, we use the multi-hop QA dataset HotpotQA (the full wiki setting) since the ground-truth of evidence passages are provided. Experiments show that our method significantly improves the accuracy of complementary evidence selection.

%%%%%%%%%%%%%%%%%%%%%%%%%%
%%%  Proposed Methods  %%%
%%%%%%%%%%%%%%%%%%%%%%%%%%

\section{Proposed Method}

\subsection{Task Definition}
Given a question ${q}$ and a mixture set of paragraphs $\mathcal{P}= \mathcal{P}^+ \cup \mathcal{P}^-$ with some paragraphs ${p} \in \mathcal{P}^+$ relevant to ${q}$ and some ${p} \in \mathcal{P}^-$ irrelevant. Our goal is to select a small subset of paragraphs $\mathcal{P}_{sel} \subset \mathcal{P}$, such that every ${p} \in \mathcal{P}_{sel}$ satisfies ${p} \in \mathcal{P}^+$ (relevancy), and all ${p} \in \mathcal{P}_{sel}$ can jointly cover all the information asked by $q$ (complementary). The off-the-shelf models select relevant paragraphs independently, thus usually cannot deal with the complementary property.  The inner dependency among the selected $\mathcal{P}_{sel}$ needs to be considered, which will be modeled in the remaining of the section.

\subsection{Model and Training}
\label{ssec:model}

\paragraph{Vector Space Modeling}
%%%%%%%%%%%%%%%%%%%%%%%%%%%%% version 1 %%%%%%%%%%%%%%%%%%%%%%%%%%%%%
We apply BERT model to estimate the likelihood of a paragraph $p$ being the supporting evidence to the question $q$, denoted as $P({p}|{q})$. Let $q$ and $p_i$ denote the input texts of a question and a passage. We feed $q$ and the concatenation of $q$ and $p_i$ into the BERT model, and use the hidden states of the last layer to represent $q$ and $p_i$ in vector space, denoted as $\bm{q}$ and $\bm{p_i}$ respectively. A fully connected layer $f(\cdot)$ followed by sigmoid activation is added to the end of the BERT model, and outputs a scalar $P({p_i}|{q})$ to estimate how relevant the paragraph $p_i$ is to the question. Note that in our implementation $\bm p_i$ is based on both $q$ and $p_i$, but we omit the condition on $q$ for simplicity.

\paragraph{Complementary Conditions}
Previous works extract evidence paragraphs according to $P({p}|{q})$, which is estimated on each passage separately without considering the dependency among selected paragraphs. To extract complementary evidence, we propose that the selected passages $\mathcal{P}_{sel}$ should satisfy the following conditions that intuitively encourage each selected passage to be a basis to support the question:

\vspace*{2mm}
\noindent $\bullet$ \textbf{Relevancy:} $\mathcal{P}_{sel}$ should have a high probability of $\sum_{p_i \in \mathcal{P}_{sel}} P(p_i|q)$;

% \vspace*{0.05in}
% \noindent $\bullet$ \textbf{Coverage:} $\mathcal{P}_{sel}$ should be able to cover all the facts asked by the question, i.e., the joint of passages in $\mathcal{P}_{sel}$ should have high similarity to $q$. For example, maximizing $cos(\sum_{i\in\{i|p_i\in\mathcal{P}_{sel}\}}\bm p_i,\bm q)$;

\vspace*{2mm}
\noindent $\bullet$ \textbf{Diversity:} $\mathcal{P}_{sel}$ should cover passages as diverse as possible, which can be measured by the average distance between any pairs in $\mathcal{P}_{sel}$, e.g., maximizing $\sum_{i,j \in \{i,j|p_i,p_j\in\mathcal{P}_{sel},i \neq j\}} \ell_1(\bm p_i,\bm p_j)$. Here $\ell_1(\cdot, \cdot)$ denotes $L_1$ distance;

\vspace*{2mm}
\noindent $\bullet$ \textbf{Compactness:} $\mathcal{P}_{sel}$ should optimize the aforementioned conditions while the size being minimal. In this work we constrain the compactness by fixing $|\mathcal{P}_{sel}|$ and meanwhile maximizing $cos(\sum_{i\in\{i|p_i\in\mathcal{P}_{sel}\}}\bm p_i,\bm q)$. We use $\cos(\cdot, \cdot)$ to encourage the collection of evidence covers what needed by the question.

\paragraph{Training with Complementary Regularization}
We propose a new supervised training objective to learn the BERT encoder for QA that optimizes the previous conditions. Note that in this work we assume a set of labeled training examples are available, i.e., the ground truth annotations contain complementary supporting paragraphs. Recently there was a growing in such datasets~\cite{yang2018hotpotqa,yao2019docred}, due to the increasing interest in model explainability.  Also, such supervision signals can also be obtained with distant supervision.

For each training instance $(q, \mathcal{P})$, we define 
\begin{align}
\small
    &\{\bm{p}_i\}^+ = \{\bm{p}_i\}, \quad\forall i\in\{i | p_i \in \mathcal{P}^+\}    \\
    &\{\bm{p}_i\}^- = \{\bm{p}_i\}, \quad\forall i\in\{i|p_i \in \mathcal{P}^-\}      \\
    &\{\bm{p}_i\}   = \{\bm{p}_i\}^+ \cup  \{\bm{p}_i\}^-
\end{align}
Denoting $y_{p_i}=1$ if ${p}_i \in \mathcal{P}^+$ and $y_{p_i}=0$ if ${p}_i \in \mathcal{P}^-$, we have the following training objective function:
%%%%% Version 1 %%%%%
% For each training instance $(q, \mathcal{P})$, we define $\{\bm{p}_i\}^+ = \{\bm{p}_i\}, \forall i\in\{i | p_i \in \mathcal{P}^+\}$; $\{\bm{p}_i\}^- = \{\bm{p}_i\}, \forall i\in\{i|p_i \in \mathcal{P}^-\}$; and $\{\bm{p}_i\} = \{\bm{p}_i\}^+ \cup  \{\bm{p}_i\}^-$. Denoting $y_{p_i}=1$ if ${p}_i \in \mathcal{P}^+$ and $y_{p_i}=0$ if ${p}_i \in \mathcal{P}^-$, we have the following training objective function:
\begin{equation} \label{eq:loss_function}
\small
\begin{aligned}
    \mathcal{L}(\{\bm{p}_i\}; \bm{q}; y) &= 
    \mathcal{L}_{sup}(\{\bm{p}_i\}; \bm{q}; y)      \\
    &+ \alpha \mathcal{L}_d(\{\bm{p}_i\}^+)         
    + \beta \mathcal{L}_c(\{\bm{p}_i\}; \bm{q}; y )
\end{aligned}
\end{equation}
where
\begin{equation}
\small
    \label{eq:loss_function_1}
    \mathcal{L}_{sup}(\{\bm{p}_i\}; \bm{q}; y) 
        = -\sum_{i} y_{p_i}\log(f(\bm{p}_i)),
\end{equation}
\begin{equation}
\small
    \label{eq:loss_function_2}
    \mathcal{L}_d(\{\bm{p}_i\}^+) 
        =  \sum_{\bm{p_i}, \bm{p_j}, i\neq j} (1- \ell_1 (\bm{p_i}, \bm{p_j})).
\end{equation}
\begin{equation}
\small
    \label{eq:loss_function_3}
    \mathcal{L}_c(\{\bm{p}_i\}; \bm{q}; y) 
        = 
        \begin{cases}
            1 - \cos (\bm{q}, \sum_{i} \bm{p}_i), \\
            \qquad \textrm{if} \quad \Pi_{p_i} y_{p_i} = 1 \\
            \max (0, \cos (\bm{q}, \sum_{i} \bm{p}_i) - \gamma), \\
            \qquad \textrm{if} \quad \Pi_{p_i} y_{p_i} = 0 
        \end{cases}
\end{equation}
where $\alpha$ and $\beta$ are the hyperparameter weights and $\ell_1(\cdot, \cdot)$ denotes L1 loss between two input vectors. Eq~\ref{eq:loss_function_1} is the cross-entropy loss corresponding to relevance condition; Eq~\ref{eq:loss_function_2} regularizes the diversity condition; Eq~\ref{eq:loss_function_3} is the cosine-embedding loss\footnote{Refer to CosineEmbeddingLoss in PyTorch.} for the compactness condition and $\gamma > 0$ is the margin to encourage data samples with better question coverage.

\subsection{Inference via Beam Search}
\paragraph{Score Function}
During inference, we use the following score function to find the best paragraph combination:
\begin{equation}
\small{
    \label{eq:score_function}
    \begin{aligned}
        g(\mathcal{P}_{sel};{q};\{\bm{p}_i\}) 
        &= 
        \sum_{p_i} P(p_i | q) + \alpha \cos(\sum_{\bm{p_i}} \bm{p}_i, \bm{q} ) \\
        &+ \beta \sum_{\bm{p_i}, \bm{p_j}, i\neq j} \ell_1 (\bm{p}_i, \bm{p}_j) 
    \end{aligned}
}
\end{equation}
where $\alpha$ and $\beta$ are hyperparameters similar to  Eq~\ref{eq:loss_function}. Note that our approach requires to encode each passage in $\mathcal{P}$ only once for each question, resulting in an $O(K)$ time complexity of encoding ($K=\vert \mathcal{P}\vert$); and the subset selection is performed in the vector space, which is much more efficient than selecting subsets before encoding.  
%%%%% Version 1 %%%%%
% where $\alpha$ and $\beta$ are hyperparameters similar to  Eq~\ref{eq:loss_function}. Note that our approach only requires to encode each passage in $\mathcal{P}$, and the subset selection is performed only in the vector space which is much more efficient ($O(K)$ time complexity of encoding, $K=\vert \mathcal{P}\vert$).  

\paragraph{Beam Search}
In a real-world application, there is usually a large candidate set of $\mathcal{P}$, e.g., retrieved passages for $q$ via a traditional IR system. Our algorithm requires $O(K)$ time encoding, and $O(K^L)$ time scoring in vector space when ranking all the combinations in $L$ candidates. Thus when $K$ becomes large, it is still inefficient even when $L=2$. We resort to beam search to deal with scenarios with large $K$s. The details can be found in Appendix~\ref{app:beam_search}.

%%%%%%%%%%%%%%%%%%%%%%%%%%
%%%     Experiments    %%%
%%%%%%%%%%%%%%%%%%%%%%%%%%

\section{Experiments}

\subsection{Settings}

\paragraph{Datasets}
Considering the prerequisite of sentence-level evidence annotations, we evaluate our approach on two datasets, a synthetic dataset \textbf{MNLI-12} and a real application \textbf{HotpotQA-50}. Data sampling is detailed in Appendix~\ref{app:data_sampling}.

\vspace{2mm}
$\bullet$ \textbf{MNLI-12} is constructed based on the textual entailment dataset MNLI \cite{N18-1101}, in order to verify the ability of our method in finding complementary evidence. In original MNLI, each premise sentence corresponds to three hypotheses sentences: entailment, neutral and contradiction. To generate complementary pairs for each premise sentence, we split each hypothesis sentence into two segments.  The goal is to find the segment combination that entails premise sentence, and our dataset, by definition, ensures that only the combination of two segments from the entailment hypothesis can entail the premise, not any of its subset or other combinations. The original train/dev/test splits from MNLI are used.
%%%%%  Version 1  %%%%%
% \noindent $\bullet$ \textbf{MNLI-12} is constructed based on the textual entailment dataset MNLI \cite{N18-1101}, in order to verify the ability of our method in finding complementary evidence. In original MNLI, each premise sentence corresponds to one entailment, one neutral and one contradiction hypothesis sentences. We take the premise as $q$, and split each of its corresponding hypotheses into two segments with a random split point near the middle of the sentence. It results in a total of 6 segments. Combining with the 6 segments corresponding to another premise, we have $\mathcal{P}$ with 12 segments for each $q$, among which the 2 segments from the original entailment sentence are annotated as $\mathcal{P}^+$. The goal is to find the segment combination that entails $q$, and our dataset, by definition, ensures that only the combination of $\mathcal{P}^+$ can entail $q$, not any of its subset or other combinations. The original train/dev/test splits from MNLI are used.

\vspace{2mm}
$\bullet$ \textbf{HotpotQA-50} is based on the open-domain setting of the multi-hop QA benchmark HotpotQA \cite{yang2018hotpotqa}. The original task requires to find evidence passages from abstract paragraphs of all Wikipedia pages to support a multi-hop question. For each $q$, we collect 50 relevant passages based on bigram BM25~\cite{godbole2019multi}. Two positive evidence passages to each question are provided by human annotators as the ground truth. Note that there is no guarantee that $\mathcal{P}_{50}$ covers both evidence passages here. We use the original development set from HotpotQA as our test set and randomly split a subset from the original training set as our development set.

% \noindent$\bullet$\textbf{HotpotQA-50} is an advanced multi-hop QA dataset customized from HotpotQA \cite{yang2018hotpotqa} with each sample consisting of one question-answer (Q-A) pair and 50 context passages. Each passage contains several sentences and any one with one or more grounded supporting evidence sentences is considered as a supporting passage. Since the 50 corresponding context passages were selected from the full English Wikipedia via our own BM25 algorithm, thus there is no guarantee to have two ground truth supporting passages for each question in the dataset.

\paragraph{Baseline}
We compare with the BERT passage ranker~\cite{nie2019revealing} that is commonly used on open-domain QA including HotpotQA. The baseline uses the same BERT architecture as our approach described in Section~\ref{ssec:model}, but is trained with only the relevancy loss (Eq~\ref{eq:loss_function_1}) and therefore only consider the relevancy when selecting evidence.

We also compare the DRN model from~\cite{harel2019learning} which is designed for the SRD task. Their ensemble system first finds the most relevant evidence to the given question, and then select the second diverse evidence using their score function. The major differences from our method are that (1) they train two separate models for evidence selection; (2) they do not consider the compactness among the evidences. It is worth mentioning that we replace their LSTM encoder with BERT encoder for fair comparison.

\paragraph{Metric}
During the evaluation we make each method output its top 2 ranked results\footnote{There is only one positive pair of evidences for each question.} (i.e. the top 1 ranked pair from our method) as the prediction. The final performance is evaluated by exact match (EM), i.e., whether both true evidence passages are covered, and the F1 score on the test sets.

\subsection{Results}

\begin{table}[t!]%[width=\textwidth]
    \small
    \centering
    \begin{tabular}{lcccc} 
        \toprule
        \multirow{2}{*}{\bf System} & \multicolumn{2}{c}{\bf HotpotQA-50} & \multicolumn{2}{c}{\bf MNLI-12}    \\
        % \cline{2-5}
         & \bf EM & \bf F1 & \bf EM & \bf F1     \\
        \midrule
        % BERT + DRN      & 10.84     & 55.42     & 6.20      & 46.07     \\
        Baseline Ranker & 16.67     & 41.29     & 41.61     & 67.57     \\
        DRN + BERT      & 1.03      & 35.37     & 6.20      & 46.07     \\
        Our Method      & \textbf{20.15}     & \textbf{49.10}     & \textbf{53.81}     & \textbf{73.18}     \\
        Upper-Bound     & 35.49     & 61.08     & 100.00    & 100.00    \\
        \bottomrule
    \end{tabular}
    \caption{Model Evaluation (\%). The upper-bound indicates the amount of true evidences contained by all candidate passages. The baseline ranker is a BERT ranker trained only with relevancy loss.}
    \label{tab:model_performance}
\end{table}

In the experiments, we have $M=3$, $N=4$ for MNLI-12 and $M=4$, $N=5$ for HotpotQA-50 with our method. The values are selected according to development performance. We follow the settings and hyperparameters used in~\cite{harel2019learning} for the DRN model. Table~\ref{tab:model_performance} shows the performance. The upper-bound measures how many pieces of true evidences enclosed by the complete set of candidate passages where our proposed ranker selects from. For HotpotQA dataset, we use a bi-gram BM25 ranker to collect top 50 relevant passages and build the basis for the experiments\footnote{This is the standard setting that starts with BM25 retrieval to make the inference time efficient enough without loss of generality.}, which inevitably leads some of the true evidences to be filtered out and makes its upper-bound less than $100\%$. For the artificial MNLI-12 dataset, all the true evidences are guaranteed to be included.

Table~\ref{tab:model_performance} shows that our method achieves significant improvements on both datasets. On HotpotQA-50, all systems have low EM scores, because of the relatively low recall of the BM25 retrieval. Only $35.49\%$ of the samples in the test set contain both ground-truth evidence passages. On MNLI-12, the EM score is around $50\%$. This is mainly because the segments are usually much shorter than a paragraph, with an average length of $7$ words. Therefore it is more challenging in matching the $q$ with the $p_i$s. Specifically, both our method and the BERT baseline surpass the DRN model on all datasets and metrics, which results from our question-conditioned passage encoding approach. Our defined vector space proves beneficial to model the complementation among the evidence with respect to a given question. The ablation study of our loss function further illustrates that the diversity and the compactness terms efficiently bring additional $20\%$/$30\%$ increase in EM score on two datasets and consequently raise the F1 score by about $8$/$6$ absolute points.
%%%%% Version 1 %%%%%
% Table~\ref{tab:model_performance} shows that our method achieves significant improvements on both datasets. On HotpotQA-50, both systems have low EM scores, because of the relatively low recall of the BM25 retrieval. Only $35.49\%$ of the samples in the test set contain both ground-truth evidence passages. On MNLI-12, the EM score is around $50\%$. This is mainly because the segments are usually much shorter than a paragraph, with an average length of $7$ words. Therefore it is more challenging in matching the $q$ with the $p_i$s.

\begin{figure}[!t]
    \centering
    \includegraphics[width=0.47\textwidth]{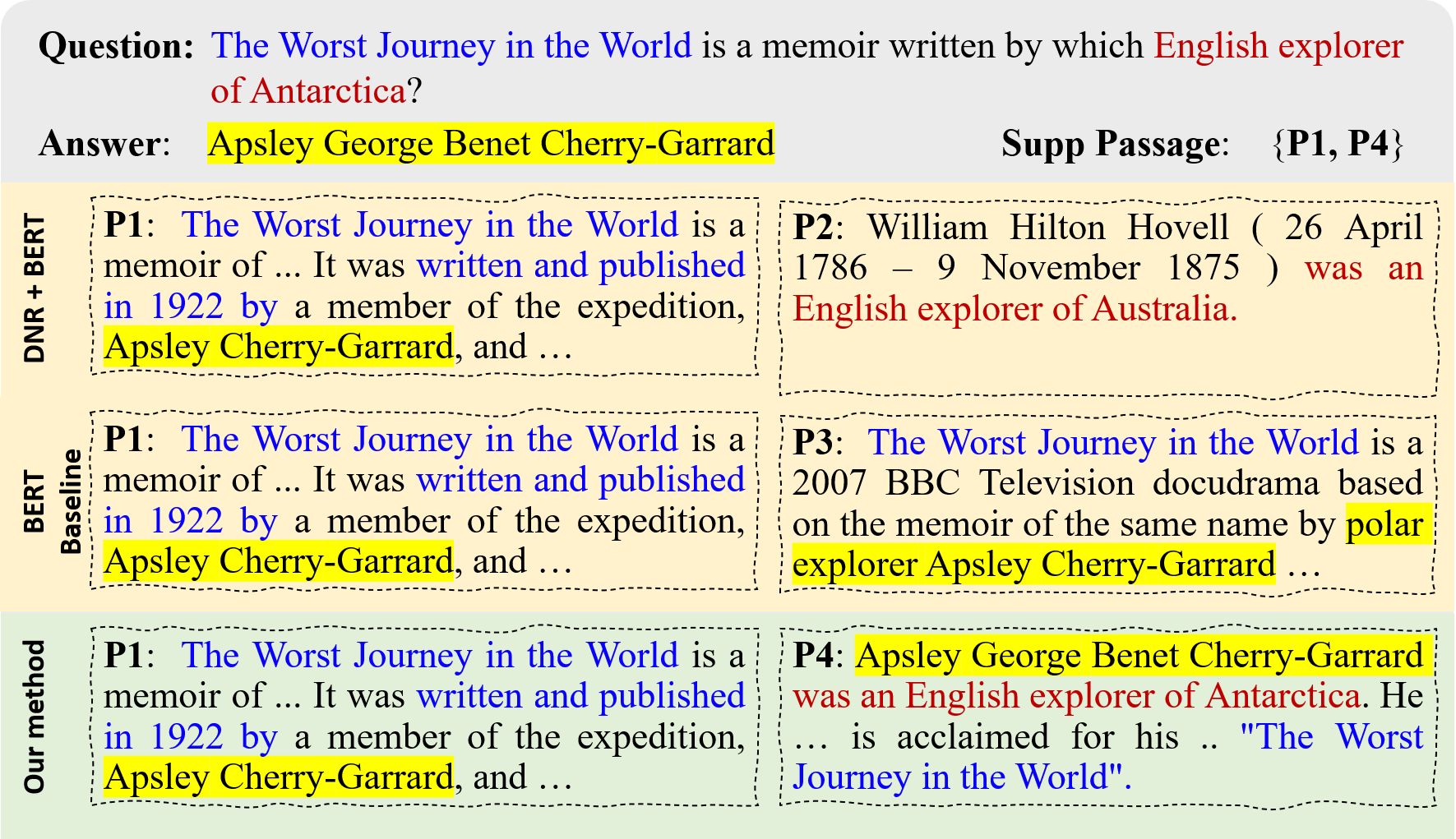}
    \includegraphics[width=0.47\textwidth]{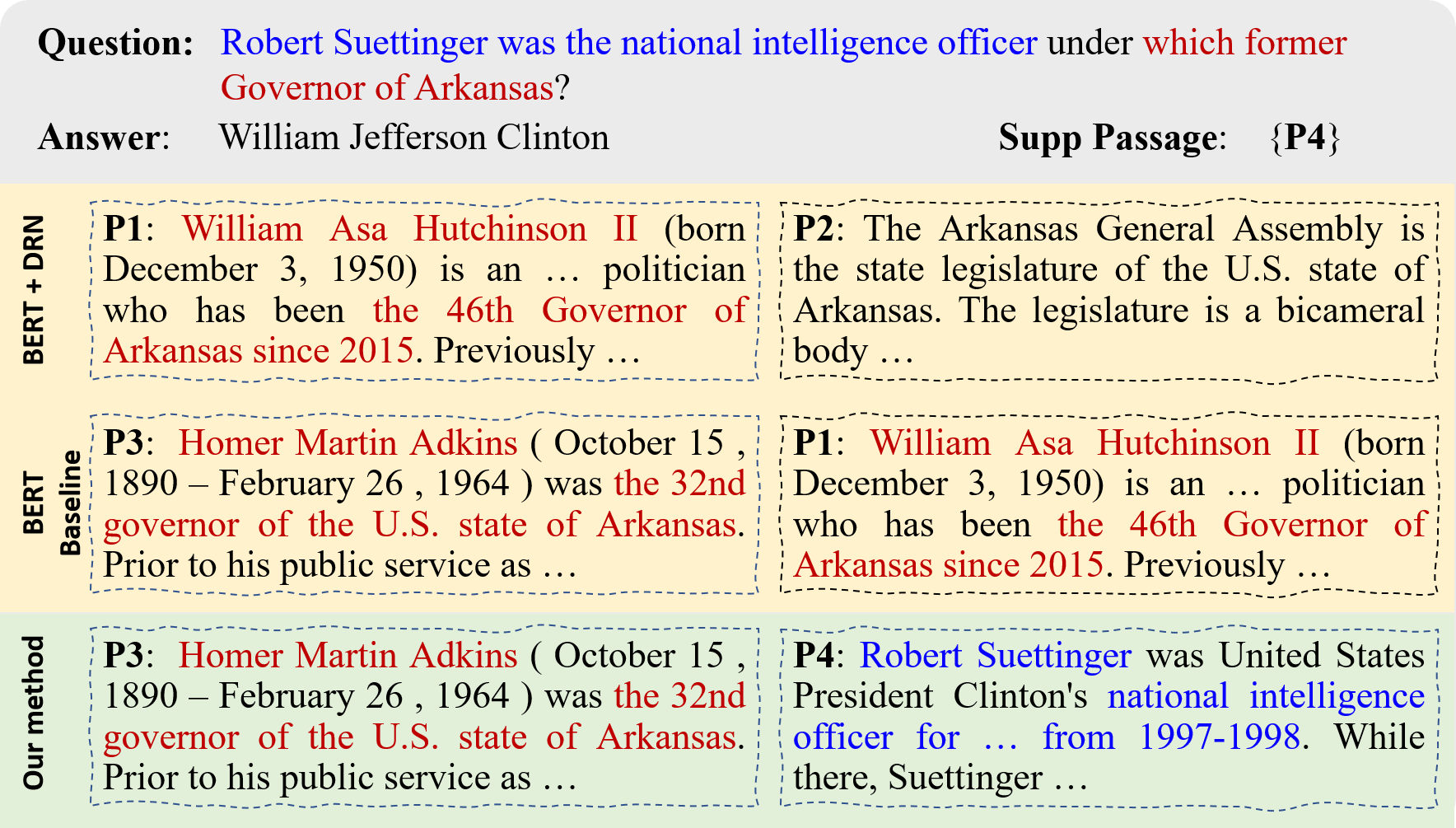}
    \caption{Gain from complementary selection. In both examples, the DRN baseline first finds the most relevant evidence to the question (left) and then select a diverse one (right); the BERT baseline model selected the top-2 most relevant passages (\textbf{P1}, \textbf{P2}) to the question regardless of their complementation; whereas our model made the selection (\textbf{P1}, \textbf{P4}) with consideration of both relevance and evidence sufficiency. Note that, in the bottom example, one of the ground-truth supporting passages and the answer were excluded when building the dataset.}
    \label{fig:result_pos}
\end{figure}

Figure~\ref{fig:result_pos} gives examples about how our model improves over the baseline. Our method can successfully select complementary passages while the baselines only select passages that look similar to the question. A more interesting example is given at the bottom where the top-50 only covers one supporting passage. The BERT baseline selects two incorrect passages that cover identical part of facts required by the question and similarly the DRN baseline select a relevant evidence and an irrelevant evidence, while our method scores lower the second passage that does not bring new information, and reaches a supporting selection. A similar situation contributes to the majority of improvement on one-supporting-evidence data sample in HotpotQA-50.

\paragraph{Inference Speed}
Our beam search with score function brings slight overheads to the running time. On HotpotQA-50, it takes 1,990 milliseconds (ms) on average to obtain the embeddings of all passages for one data sample whereas our vector-based complementary selection only adds an extra 2 ms which can be negligible compared to the encoding time.  

% Time-wise, our beam search with score function brought additional overheads to the running time. Specifically for every 1000 runs, our score function added an extra 87 milliseconds (ms) to the baseline of 46 ms (Only inclusive of ranking step). The beam search algorithm made the overhead $M \times N$ times larger. For example, in our experiment on MNLI-12 with $M=3$ and $N=4$, the inference time of our method for 1918 test samples was 1997 ms on average against 88 ms by the baseline. It was considered acceptable compared with the benefit it brought.

\subsection{Future Work}
The latest dense retrieval methods~\cite{lee2019latent,karpukhin2020dense,guu2020realm} show promising results on efficient inference on the full set of Wikipedia articles, which allows to skip the initial standard BM25 retrieval and avoid the significant loss during the pre-processing step. Our proposed approach is able to directly cooperate with these methods as we all work in the vector space. Therefore, the extension to dense retrieval can be naturally the next step of our work.

%%%%%%%%%%%%%%%%%%%%%%%%%%
%%%     Conclusion     %%%
%%%%%%%%%%%%%%%%%%%%%%%%%%

\section{Conclusion}
In the paper, we propose a new problem of complementary evidence identification and define the criterion of complementary evidence in vector space. We further design an algorithm and a loss function to support efficient training and inference for complementary evidence selection. Compared to the baseline, our approach improves more than $20\%$ and remains to scale well to the computationally complex cases.

%%%%%%%%%%%%%%%%%%%%%%%%%%
%%%   Acknowledgment   %%%
%%%%%%%%%%%%%%%%%%%%%%%%%%

\section*{Acknowledgment}
A special thank to Rensselaer-IBM Artificial Intelligence Research Collaboration (RPI-AIRC) for providing externship and other supports. This work is funded by Cognitive and Immersive Systems Lab (CISL), a collaboration between IBM and RPI, and also a center in IBM’s AI Horizons Network. 
% (Dr. Lisa Amini and Dr. Pin-Yu Chen) 

\bibliographystyle{acl_natbib}
\bibliography{reference}

\clearpage
\onecolumn

\appendix
\small

\section{Complementary Evidence Selection via Beam Search}
\label{app:beam_search}
For efficient inference when $L = 2$, we start to select the top-$N$ ($N \ll K$) most relevant passages. Then we score the combinations between each passage pair in the top-$N$ set and another top-$M$ set.  This reduces the complexity from $O(K^2)$ to $O(MN)$.  $M$ is a hyperparameter corresponding to the beam size. In a more general setting with $L \ge 2$, we have an algorithm with the complexity of $O((L-1)MN)$ instead of $O(K^L)$, which is shown in algorithm \ref{alg:beam}. 

\begin{algorithm}[H]
\label{alg:beam}
\SetAlgoLined
\KwData{Vector representation of question ($\bm{q}$), vector representation of all the $N$ passages $\{p_n\}$ ($\{\bm{p_n}\}$); the maximum number of passage to select ($L$); the beam size ($M$); a vector of weights for all regularization terms $\bm{\lambda}$.}
\KwResult{The top ranked complementary passages.}
 \tcc{Predict the probability $P(\bm{p_i})$ of being a supporting passage for each passage $\bm{p_i}$ given $\bm{q}$}
\For{$i \in [1, N]$} {
    $P(\bm{p_i}) \gets f(\bm{q}, \bm{p_i})$\;
}
Rank the passages by $P(\bm{p_i})$\;
$\bm{P_{span}}$ = $[ ]$ \\
Pick $M$ passages with top $P(\bm{p_i})$ into $\bm{P_{span}}$\;
\For{$depth \in [2, L]$} 
{
    $\bm{P'_{span}}$ = $[ ]$ \;
    \For{$j \in [1, M]$}
    {
        \tcc{$\bm{P_j}$ is a selected subset, $s_j$ is the corresponding score}
        Pop the $j$-th tuple $(\bm{P_j}, s_j)$ from $\bm{P_{span}}$\;
        \For{$n \in [1, N]$} 
        {
            \If{The set $\bm{P_j}\cup \{\bm{p_n}\}$ is covered by $\bm{P'_{span}}$}{continue}
            \tcc{$\bm{r}_{n}$ is the regulation increases by adding $\bm{p_n}$ to $\bm{P_j}$}
            Put $(\bm{P_j}\cup \{\bm{p_n}\}, s_j + P(\bm{p}_n) + \bm{\lambda} \bm{r}_{n})$ into $\bm{P'_{span}}$\;
            \If{More than $M$ tuples added based on $\bm{P_j}$}{break}
        }
    }
    Rank $\bm{P'_{span}}$ according to the scores\;
    $\bm{P_{span}} \gets \bm{P'_{span}}[1:M]$
}
\textbf{Return} $\bm{P_{span}}$[0]
\caption{Complementary Evidence Selection via Beam Search}
\end{algorithm}

\section{Data Sampling}
\label{app:data_sampling}

\paragraph{MNLI-12}
In original MNLI, each premise sentence $P$ corresponds to one entailment $E_P$, one neutral $N_P$ and one contradiction $C_P$. We take the premise $P$ as $q$, and split each of its corresponding hypotheses into two segments with a random cutting point near the middle of the sentence, resulting in a total of 6 segments $\{E_P^1, E_P^2, N_P^1, N_P^2, C_P^1, C_P^2\}$. Mixing them with the 6 segments corresponding to another premise $X$, we can finally have $\mathcal{P}^+ = \{E_P^1, E_P^2\}$ and $\mathcal{P}^- = \{N_P^1, N_P^2, C_P^1, C_P^2, E_X^1, E_X^2, N_X^1, N_X^2, C_X^1, C_X^2\}$. Consequently, we sample one positive and eight negative pairs respectively from $\mathcal{P}^+$ and $\mathcal{P}^-$. A pair like $\{E_P^1, C_X^2\}$ is considered as negative. To ensure the segments are literally meaningful, each segment is guaranteed to be longer than 5 words.

\paragraph{HotpotQA}
In HotpotQA, the true supporting paragraphs of each question $q$ are given. Therefore, we can easily form $\mathcal{P}^+$ and $\mathcal{P}^-$ and sample positive and negative pairs of paragraphs respectively from $\mathcal{P}^+$ and $\mathcal{P}^-$. A special pair that contains one true supporting paragraph and one non-supporting paragraph is considered as a negative pair.

\clearpage
\twocolumn

\end{document}